\documentclass[letterpaper, 10 pt, conference]{ieeeconf}  

\IEEEoverridecommandlockouts                              
\usepackage{cite}
\usepackage{graphicx} 
\usepackage{amsmath}
\usepackage{amssymb}
\usepackage{color}
\usepackage[font=small]{caption}
\usepackage{booktabs}
\usepackage{subcaption}
\usepackage{multirow}

\usepackage{algorithm}
\usepackage{algorithmic}
\usepackage[algo2e]{algorithm2e}

\title{\LARGE \bf
Distributed Deep Reinforcement Learning for Intelligent \\ Traffic Monitoring with a Team of Aerial Robots
}

\author{Behzad Khamidehi and Elvino S. Sousa\\
\thanks{The authors are with the Department of Electrical and Computer Engineering, University of Toronto, ON M5S 1A1, Canada 
{ \tt\small  \noindent \{b.khamidehi, es.sousa\}@utoronto.ca}}
}

\begin{document}

\maketitle
\thispagestyle{empty}
\pagestyle{empty}

\begin{abstract}
This paper studies the traffic monitoring problem in a road network using a team of aerial robots. The problem is challenging due to two main reasons. First, the traffic events are stochastic, both temporally and spatially. Second, the problem has a non-homogeneous structure as the traffic events arrive at different locations of the road network at different rates. Accordingly, some locations require more visits by the robots compared to other locations. To address these issues, we define an uncertainty metric for each location of the road network and formulate a path planning problem for the aerial robots to minimize the network's average uncertainty. We express this problem as a partially observable Markov decision process (POMDP) and propose a distributed and scalable algorithm based on deep reinforcement learning to solve it. We consider two different scenarios depending on the communication mode between the agents (aerial robots) and the traffic management center (TMC). The first scenario assumes that the agents continuously communicate with the TMC to send/receive real-time information about the traffic events. Hence, the agents have global and real-time knowledge of the environment. However, in the second scenario, we consider a challenging setting where the observation of the aerial robots is partial and limited to their sensing ranges. Moreover, in contrast to the first scenario, the information exchange between the aerial robots and the TMC is restricted to specific time instances. We evaluate the performance of our proposed algorithm in both scenarios for a real road network topology and demonstrate its functionality in a traffic monitoring system.

\end{abstract}

\section{INTRODUCTION}

Unmanned aerial vehicles (UAVs) have recently attracted considerable interest in a wide range of applications. Aerial reach, high mobility, and cost-effective deployment are the key features that make the UAVs an ideal candidate for applications such as drone delivery \cite{choudhury_stanford,TITS_Synchronized, khamidehi2021dynamic}, search and rescue \cite{ICAPS2015, IEEE_Access}, wireless communications \cite{zeng2016wireless,TITS_2020content,Zhang_5G}, and mapping, tracking, and monitoring \cite{IROS_wilefire, IROS_firefighting_UAVs, IROS_Cinematography}. UAV-assisted traffic monitoring in urban areas is another emerging application that can play a key role in intelligent transportation systems (ITSs) \cite{Guvenj_comm_mag}. Currently, the monitoring is performed by a set of networked cameras installed in different locations of the road network. However, the implementation cost of these systems is usually high. Hence, they are not economical solutions for monitoring short-term traffic events. Moreover, these systems do not offer flexible solutions for the dead zones or locations without appropriate infrastructures \cite{VTC_UAV_traffic}. To overcome these limitations, we can integrate UAVs into traffic monitoring systems.

The UAV-assisted traffic monitoring has been investigated in several recent studies \cite{VTC_UAV_traffic, IV_cooperative2020, TITS_parameter_estimation, zhu2018urban, TITS_Bidirectional, TITS_trajectory_extraction, IROS_parking, VTC_optimizing_tour, ITSC_TSP, Aerospace, WCNC_monitoring}. In \cite{VTC_UAV_traffic}, a single UAV traffic monitoring system has been developed to capture traffic videos from the road network and send them to the traffic management center (TMC). In \cite{IV_cooperative2020}, a cooperative traffic monitoring system has been considered to help terrestrial vehicles to have full information about their surroundings based on the UAV's images. In \cite{TITS_parameter_estimation, zhu2018urban, TITS_Bidirectional}, the authors adopted deep learning to estimate the traffic flow parameters from the UAV's captured videos. In \cite{IROS_parking}, the authors developed a parking occupancy detection algorithm based on the UAVs' images. In \cite{VTC_optimizing_tour}, multi-UAV tour planning problem has been studied to monitor the traffic on a given road network. However, the proposed algorithm is an offline planning one that only uses the road network topology and does not consider any dynamics in the system. In \cite{ITSC_TSP}, an extended multiple traveling salesman problem has been studied to schedule a team of UAVs for traffic monitoring purposes. However, similar to \cite{VTC_optimizing_tour}, the considered problem is offline, where the visit points and the corresponding visit time windows are known.  
In \cite{WCNC_monitoring}, a dynamic traffic monitoring problem has been investigated where the authors assumed the UAVs can accurately detect the vehicles and estimate their true positions. Given this information, a simple path planning algorithm has been proposed to follow the gravity center of the vehicle clusters in the road network. 

\begin{figure}[t]
	\centering
	\includegraphics[width=0.8\columnwidth]{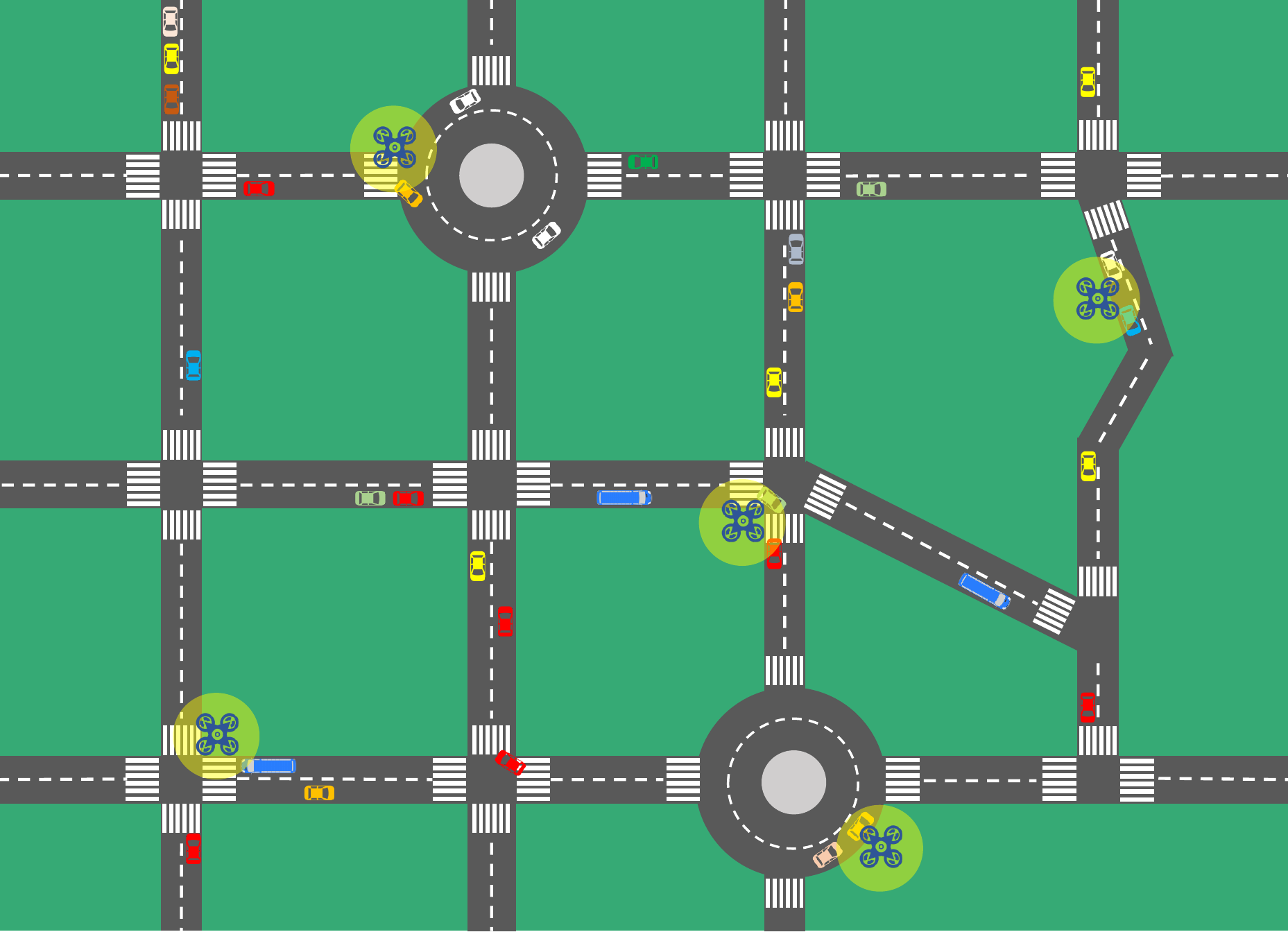}
	\caption{Traffic monitoring using a network of aerial robots.}
	\label{fig:model}
\end{figure}

\noindent \textbf{Our Contributions.} 
In contrast to the mentioned studies that focus on either an offline problem setting \cite{VTC_optimizing_tour, ITSC_TSP}, or a scenario with perfect knowledge of the road vehicles \cite{WCNC_monitoring}, we consider a dynamic and online problem setting with partial observations, and solve the navigation problem for a team of UAVs under this limitation. Due to the random and time-varying nature of the traffic events in the road networks, the UAVs must regularly visit different locations to catch the traffic events. To address this issue, we define an \textit{uncertainty} metric for each location of the road network and formulate a path planning problem for the UAVs to minimize the network's average uncertainty. We express this multi-UAV traffic monitoring problem as a partially observable Markov decision process (POMDP) and propose a decentralized and scalable solution based on deep reinforcement learning (RL) to solve the problem. Depending on the communication mode between the agents and the TMC, we consider two scenarios for the traffic monitoring problem. In the first scenario, we assume that the agents (UAVs) continuously communicate with the TMC and hence, they have perfect and real-time knowledge of the environment. However, in the second scenario, we consider a challenging setting where the information exchange between the agents and the TMC is restricted to specific time instances. In other words, we do not consider a continuous communication between the agents and the TMC. Moreover, we assume that the visibility of each UAV is limited to its sensing range, and hence, it has partial observation from its surrounding environment. We evaluate the performance of our proposed method for a real road network topology in downtown Toronto. Evaluation results show the effectiveness of our proposed algorithm for traffic monitoring purposes.

\section{SYSTEM MODEL}

We consider a team of $N$ aerial vehicles that monitor traffic conditions in a given road network, as shown in Fig. \ref{fig:model}. We use a grid-world representation of size $M \times M$ for the environment. The total number of grid-cells is represented by $K \triangleq M\times M$ and index $k$ is used to refer to the $k$-th cell. The task of the UAVs is to visit different locations of the road network to capture images from the traffic conditions. These images will be sent to the TMC for traffic regulation purposes.

\subsection{Agent Model}
We use index $i$ to refer to the $i$-th agent (UAV). The position of the $i$-th agent at time $t$ is represented by $\mathbf{p}_i (t)$. Each UAV has a downward-facing camera that captures images from the streets and the traffic conditions. We assume that the camera's field of view (FoV) can cover one grid cell (currently positioned cell). The UAV also has radio sensors (transmitter/receiver) to send its collected data to the TMC. The static global map of the environment is also given to all agents. Using a GPS sensor, each agent can localize itself on the map. This map also gives the locations of the static obstacles and no-fly zones. The agents must avoid collision with both these static obstacles and the moving objects, which are other agents in our model. Moreover, since the task of the aerial vehicles is to gather information from the road network, it will be a waste of resources if two agents cover the same cell simultaneously. Hence, we have 
\begin{equation}
    \label{eq:collision_free}
    \mathbf{p}_{i_1}(t)\neq \mathbf{p}_{i_2}(t), \  \forall i_1 \neq i_2, \ \forall t.
\end{equation}

\subsection{Uncertainty Model}

The goal of aerial vehicles is to monitor the traffic condition and collect information about the traffic events such as traffic jam(s), accident(s), traffic law violation(s), etc. These events can appear in different locations of the environment in a random and time-varying basis. To address the randomness of the events, we define an uncertainty metric for each location of the road network. Given this uncertainty model, we can form an uncertainty map for the environment. This map gives the probability of having an event in each location, or equivalently, it shows the locations that require a visit by the aerial vehicles because there is no confidence about their traffic conditions. Depending on the communication mode between the agents and the TMC, we consider two models for the uncertainty in our system. 

\noindent \textbf{\textit{Scenario I} (continuous communication).} In this scenario, we assume that the agents have a continuous communication with the TMC. Hence, the locations of the traffic events are given to the agents by the TMC. Let $e_k (t)$ denote an indicator function taking value of $1$ if there is an active event in the $k$-th  grid-cell at time $t$ and $0$, otherwise. In this scenario, the UAVs know the values of $e_k(t), \forall k$, at each time $t$. The goal of the UAVs is to visit the locations with active events, i.e., the locations with $e_k (t) =1$. We define the uncertainty of the $k$-th cell as
\begin{equation}
    \label{eq:uncertainty_1}
    u_k (t) = \begin{cases} 1 & \text{ if } e_k(t)=1, \\
    0 & \text{ otherwise.}
    \end{cases}
\end{equation}
To reduce the uncertainty of the environment, the agents should visit locations with $e_k(t)=1$. After visiting a location with an active event, its corresponding $e_k(t)$ will be $0$, meaning that an agent has visited the location and there is no further uncertainty about it. The value of uncertainty remains $0$ until another event emerges at this location.

\noindent \textbf{\textit{Scenario II} (limited communication).} In this case, the communication between the agents and the TMC is limited to specific time instances. As a result, the agents do not have complete information about the events and their locations. 
Let $\nu_k (t)$ denote an indicator function that takes value of $1$ if the $k$-th cell is visited by one of the agents at time $t$. Otherwise, we have $ \nu_k (t) = 0$. Moreover, let $\tau_k$ denote the last time that the $k$-th grid-cell has been visited by an agent. Under a Poisson distribution, the probability that we have at least one event in the $k$-th cell in interval $[\tau_k, t)$ is $1- e^{-\alpha_k (t-\tau_k)}$, where $\alpha_k$ is the rate of event arrival in the $k$-th grid-cell. We can use this probability as the uncertainty metric. In other words, at time $t$, we can define the uncertainty of the $k$-th cell as
\begin{equation}
    \label{eq:uncertainty_2}
    u_k (t) = 1 - e^{-\alpha_k (t-\tau_k (t))},
\end{equation}
where 
\begin{equation}
    \label{eq:last_visit_time}
    \tau_k (t) = \displaystyle \max_{0 \leq \tau' \leq t} \left \{\tau'| \ \nu_k(\tau') = 1 \right \}.
\end{equation}
According to this definition, when $t=\tau_k (t)$, the $k$-th cell is visited by an agent. Hence, there is no uncertainty about this cell, and the value of uncertainty is $0$. However, as $t$ increases, the value uncertainty increases based on \eqref{eq:uncertainty_2}. When the value of $t$ becomes sufficiently large, the uncertainty tends to $1$. This implies that there is no further confidence about the corresponding cell as it has been a long time since the last agent visited this cell.

\subsection{Sensing Range and Information Exchange}

As discussed earlier, in \textit{scenario I}, each agent has complete knowledge of the environment as the locations of active events (events with $e_k (t) = 1$) and other agents' real-time locations are given to each agent by the TMC. As a result, there is no limitation for the sensing range of each UAV in \textit{scenario I}. We can assume that each agent has access to the global and real-time uncertainty map of the environment in this scenario. However, in \textit{scenario II}, we assume that the visibility of each agent is limited to its sensing range. Let $\mathcal{N}_i(t)$ denote the set of agents that are located in the sensing range of the $i$-th agent at time $t$. We have 
\begin{equation}
    \mathcal{N}_i (t) = \left\{i' \ \big| \  \big\rVert \mathbf{p}_i (t) - \mathbf{p}_{i'} (t) \big\rVert \leq r_s, i' \neq i \right\},
\end{equation}
where $r_s$ is the sensing range of each robot. In \textit{scenario II}, at time $t$, the $i$-th agent only knows locations of the agents in $\mathcal{N}_i (t)$. Moreover, in this scenario, the information exchange between the agents and the TMC is performed every $T_u$ time units. For this purpose, each agent has a memory that keeps a record of the last $L$ locations (cells) the agent has visited. This information is sent to the TMC every time the agent and the TMC communicate (every $T_u$ time units). Using this information, the TMC updates the uncertainty map of the environment and sends it back to the agents. It is worth mentioning that there is no need for synchronous communication between all agents. In other words, the agents can communicate with the TMC at different time instances.

\subsection{Uncertainty Map Update}

In \textit{scenario I}, all agents have access to the global uncertainty map. Let $\mathcal{V}^n(t:t+1)$ and $\mathcal{V}^v(t:t+1)$ denote the set of indices corresponding to the cells that have new events in interval $[t, t+1)$ and the cells that are visited by one of the agents in interval $[t,t+1)$, respectively. To update the uncertainty map in this scenario, we set
\begin{align*}
    e_k(t+1) = 0, \ \ \ \ \forall k \in \mathcal{V}^v(t:t+1),\\
    e_k (t+1) =1, \ \ \ \ \forall k \in \mathcal{V}^n(t:t+1).
\end{align*}
For all other grid-cells, we have $e_k(t+1) = e_k(t)$. Using these values, the new uncertainty can be derived based on the uncertainty equation in \eqref{eq:uncertainty_1}.

In \textit{scenario II}, as we discussed earlier, the agents do not have access to the global uncertainty map at all time instances. Hence, each agent maintains a local uncertainty map for itself and updates this map using its local information. Once the agent communicates with the TMC, it can update its local map with the global uncertainty map (every $T_u$ time units). In what follows, we discuss how the uncertainty map is updated locally and globally by each agent and the TMC, respectively.
\begin{itemize}
    \item \textit{Local update}: In the time interval between two consecutive updates by the TMC, each UAV updates its own uncertainty map using its local collected data. In particular, at time $t$, each agent sets $\tau_k (t+1) = t+1$ for its current cell and the cells that are in its sensing range and have been visited by one of the agents in time interval $[t,t+1)$. For other cells, the agent sets $\tau_k(t+1) = \tau_k(t)$. Using the value of $\tau_k(t+1)$ and \eqref{eq:uncertainty_2}, the agent updates its local uncertainty map.
    \item \textit{Global update}: Every $T_u$ time units, the agents send their visited locations and the corresponding visit times to the TMC. Let $\mathcal{V}^v(t:t+T_u)$ denote the set of all cells that have been visited by at least one of the agents in interval $[t,t +T_u)$. We have
    \begin{equation*}
        \mathcal{V}^v(t:t+T_u) = \displaystyle \bigcup_{i=1}^{T_u} \mathcal{V}^v (t+i-1:t+i).
    \end{equation*}
    If $k \in \mathcal{V}^v (t:t+T_u)$, the TMC will set $\tau_k(t+T_u)$ to the time that the cell has been visited by an agent. In case that the $k$-th cell has been visited more than once during $[t, t+T_u)$, the TMC sets $\tau_k(t+T_u)$ to the last time that the cell has been visited. For other cells that have not been visited by any of the agents in interval $[t,t+T_u)$, the TMC sets $\tau_k(t+T_u) = \tau_k(t)$. Using the value of $\tau_k(t+T_u)$ and \eqref{eq:uncertainty_2}, the TMC evaluates the new uncertainty map and broadcasts it to the agents. 
\end{itemize}

\subsection{Problem Definition}
To formulate the problem, first, we define the average uncertainty of the environment as
\begin{equation}
\label{eq:avg_uncertainty}
\bar{u} = \frac{1}{T}\sum_{t=1}^{T} \sum_{k=1}^{K} u_k (t),
\end{equation}
where $T$ is the total monitoring time. The goal of the agents is to minimize the average uncertainty ($\bar{u}$) in the environment. To achieve this goal, the UAVs need appropriate paths to follow. 
These paths must satisfy the condition in \eqref{eq:collision_free} throughout the UAVs' flights. In the next section, we describe how to formulate the path-planning problem as a POMDP and solve it using reinforcement learning techniques. 

\section{Methodology}

\subsection{Reinforcement Learning Overview}

Reinforcement learning is a framework for solving sequential decision-making problems. In RL, the agent interacts with the environment in a sequence of discrete time instances as follows: At each time $t$, the agent receives observation $\mathbf{o}_t$ from the environment. This observation is a representation of the true state of the environment, denoted by $\mathbf{s}_t$, which is not directly observable by the agent. Using this observation, the agent takes action $a_t$, receives reward $r_{t+1}$ from the environment and goes to a new state $\mathbf{s}_{t+1}$ which is available to the agent through its observation $\mathbf{o}_{t+1}$, and this procedure continues. To formulate this interaction, we can use POMDPs. A POMDP can be expressed as a tuple $\prec \mathcal{S}, \mathcal{A}, \mathcal{T}, R, \Omega, \mathcal{O}, \gamma \succ$, where $\mathcal{S}$ is the state space, $\mathcal{A}$ is the finite action space, $\mathcal{T}(s',s,a)= P(s'|s,a)$ is the transition function that maps actions and states to a distribution over the next states, $R: \mathcal{S} \times \mathcal{A} \rightarrow \mathbb{R}$ is the reward function, $\Omega$ is the observation space, $\mathcal{O}(s,a,o)=P(o|s,a)$ is the observation function, and $\gamma \in (0,1]$ is the discount factor. The action selection mechanism of the agent is called \textit{policy} and is denoted by $\pi(a|\mathbf{o})=P(a_t = a | \mathbf{o}_t = \mathbf{o})$. Let $Q_{\pi}(\mathbf{o},a)$ denote the expected return the agent receives over the long run if it starts from a state with observation $\mathbf{o}$, take action $a$, and follow policy $\pi$ afterwards. This function is referred to as \textit{Q-function} and is defined as
\begin{equation}
    Q_{\pi} (\mathbf{o},a) = \mathbb{E}_{\pi} \left\{\sum_{k=0}^{\infty} \gamma^k r_{t+k+1} | \mathbf{o}=\mathbf{o}_t, a = a_t \right\}.
\end{equation}
The goal of the RL agent is to find a policy $\pi^*$ that maximizes $Q_{\pi} (\mathbf{o},a)$. In problems with a large state/action space, we can use a multi-layer neural network to represent Q-function. In other words, we have $Q_{\pi} (\mathbf{o}, a) \approx
Q(\mathbf{o}, a; \theta)$, where $\theta$ is the parameter of the neural network. The corresponding neural network is called \textit{Q-network}.
\vspace{0.5mm}

\noindent \textbf{Deep Q-networks (DQN).} To obtain the Q-function, we can use DQN \cite{mnih_Atari, mnih2015human}. The key components of this algorithm are the target network and the experience replay memory. We denote the parameter of the target network with $\theta^-$ which is a periodic copy of $\theta$. At each time $t$, the agent implements an $\epsilon$-greedy algorithm to explore its environment. Upon taking the action, the agent's experience tuple, i.e., $(\mathbf{o}_t, a_t, \mathbf{o}_{t+1}, r_{t+1})$, is stored in the replay memory $\mathcal{D}$. To update $\theta$, we sample a mini-batch of size $b$ from $\mathcal{D}$ and define the target values for each sample as 
\begin{equation}
    \label{Target_value}
    y_t = r_{t+1} + \gamma \max_{a'} Q(\mathbf{o}_{t+1},a'; \theta^-).
\end{equation}
By minimizing the loss function defined as
\begin{equation}
\label{eq:loss}
    \mathcal{L} (\theta) = E_{\pi} \{ (Q(\mathbf{o}_t,a_t;\theta) - y_{t})^2\},
\end{equation}
we can update $\theta$.

\subsection{Multi-robot Traffic Monitoring as a POMDP}
Now, we can formulate the problem as a POMDP. In what follows, we introduce the components of our considered POMDP.

\noindent \textbf{State.} The state of the $i$-th agent is defined as 
\begin{equation}
    \label{eq:state}
    \mathbf{s}_i (t) = [\mathbf{p}_i(t), \mathbf{M}, \mathbf{p}_{-i} (t), \mathbf{U} (t)],
\end{equation}
where $\mathbf{p}_i(t) \in \mathbb{R}^{2}$ is the position of the $i$-th robot in the map, $\mathbf{M}\in \mathbb{R}^{M \times M}$ is the map of the environment which consists of the static obstacles, streets, intersections, etc., $\mathbf{p}_{-i} (t) \in \mathbb{R}^{2 \times (N-1)}$ includes the locations of all other robots (except $i$), and $\mathbf{U}(t) \in \mathbb{R}^{M \times M}$ is the global uncertainty map of the environment.

\noindent \textbf{Observation.} In \textit{scenario I}, the observation of each agent is the same as its state. As a result, the considered POMDP reduces to a fully observable MDP. However, in \textit{scenario II}, the observation of each agent is limited to a certain range and the states are not directly observable by the agents. Therefore, at time $t$, the $i$-th agent ($\forall i$) only knows the locations of the agents that are in $\mathcal{N}_i (t)$. Accordingly, it can updates its uncertainty map only using this limited information and it does not know the global and true uncertainty of the whole road network. Let $\mathbf{U}_i (t) \in \mathbb{R}^{M \times M}$ denote the $i$-th agent's local uncertainty map at time $t$. 
This map is updated locally by the $i$-th agent. According to our discussion, the observation of the $i$-th agent has the following components
\begin{equation}
\label{eq:obs_1}
    \left[\mathbf{p}_i (t), \mathbf{M}, \left\{\mathbf{p}_{i'} (t), \ \forall i' \in \mathcal{N}_i(t) \right\}, \mathbf{U}_i (t)\right].
\end{equation}
Instead of \eqref{eq:obs_1}, we can use a multi-channel representation for the observation and define $\mathbf{o}_i(t)$ as 
\begin{equation}
    \label{eq:observation}
    \mathbf{o}_i(t) = [\mathbf{P}_i (t), \mathbf{M}, \mathbf{U}_i (t)].
\end{equation}
In this representation, $\mathbf{P}_i \in \mathbb{R}^{M \times M}$ is the position channel that encodes the position of the $i$-th agent and its neighbouring agents. For this channel, we use different values to differentiate between the ego vehicle ($i$-th agent) and those in $\mathcal{N}_i (t)$. We also use different values to differentiate between different objects of the second channel such as obstacles, roads, etc.

\begin{algorithm}[t]
        \small
        \SetAlgoLined
        \textbf{Initialization:}\\
        Initialize network $Q$ with random parameter $\theta$\\
        Initialize the target network $Q^-$ with $\theta^- = \theta$\\
        Initialize the replay memory $\mathcal{D}$.\\
        \textbf{Training}:\\
        \For{$ \text{episode}=1, 2, \ldots, E$}{
        t = 0\\
        Initialize simulator and set $\mathcal{D}_i = \varnothing, \forall i$.\\
        \While{$t < T_\text{ep}$}{
        \For{each agent $i$}{
        Observe $\mathbf{o}_i(t)$, take action $a_i(t)$ using an $\epsilon$-greedy policy, receive reward $r(t)$ and observe $\mathbf{o}_i (t+1)$.\\
        Add $(\mathbf{o}_i(t), a_i(t), r_i(t), \mathbf{o}_{i}(t+1))$ to $\mathcal{D}_i$ \\
        Update the local uncertainty $\mathbf{U}_i (t)$ using the local information.\\
        \If{$t$ mod $T_u=0$}{
        Send all the experience tuples in $\mathcal{D}_i$ to the TMC.\\
        Receive the updated uncertainty map from the TMC and update the local uncertainty map accordingly.\\
        $\mathcal{D}_i = \varnothing$.
        }
        }
        Sample a mini-batch of size $b$ from $\mathcal{D}$ and update the network parameter $\theta$. \\ 
        $t = t+1$\\
        }
        If $episode$ mod $f=0$, update the target network as $\theta^- = \theta$.
        }
        \caption{Traffic monitoring based on Distributed-DQN.}        
        \label{alg:distributedDQN}
\end{algorithm}

\begin{figure*}
     \centering
     \begin{subfigure}{.32\textwidth}
         \centering
         \includegraphics[width=0.75\textwidth]{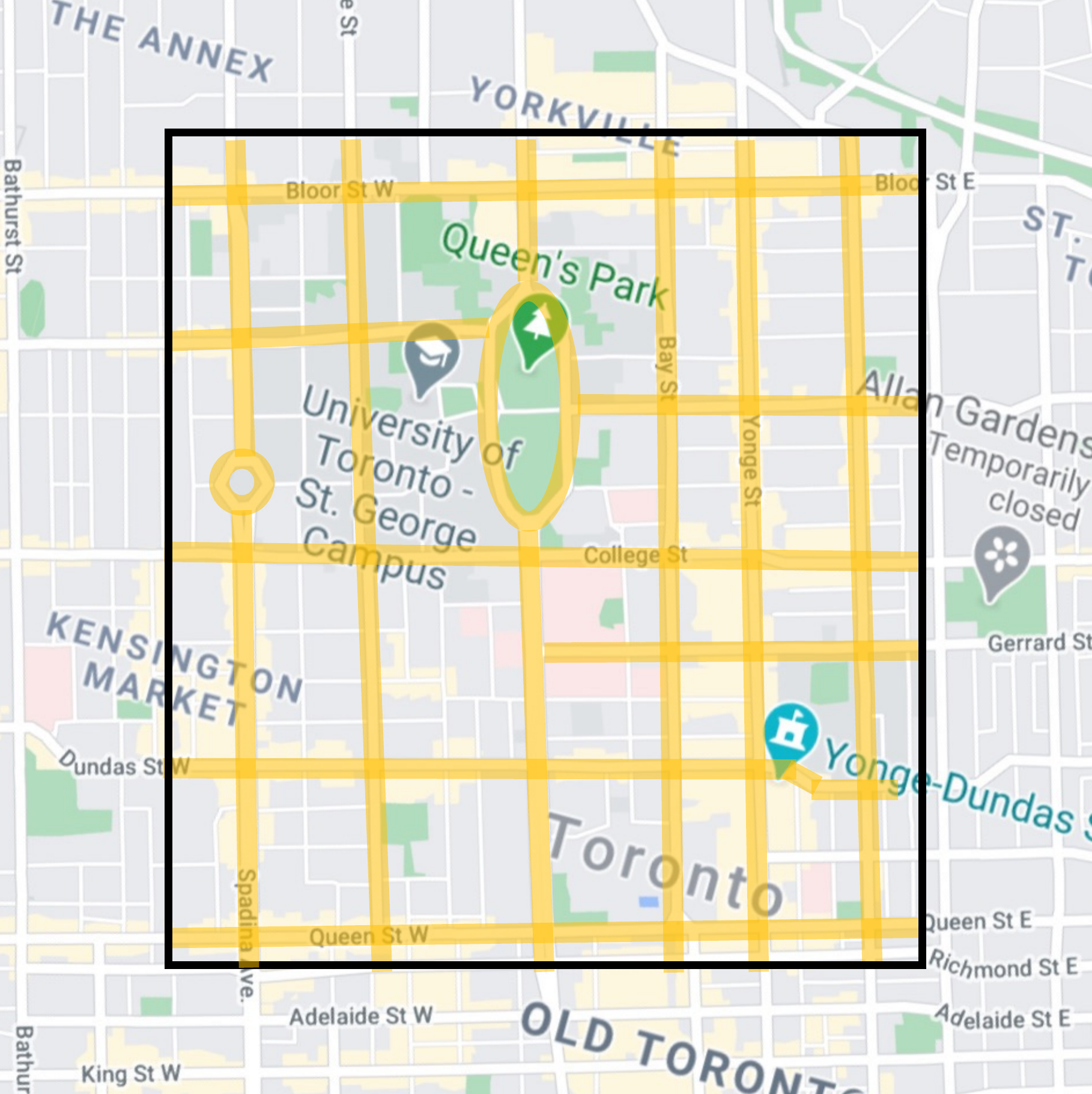}
         \vspace{0.0mm}
         \caption{}
         \vspace{-0.0mm}
         \label{fig:Toronto}
     \end{subfigure}
     \begin{subfigure}{.66\textwidth}
         \centering
         \includegraphics[width=0.95\textwidth]{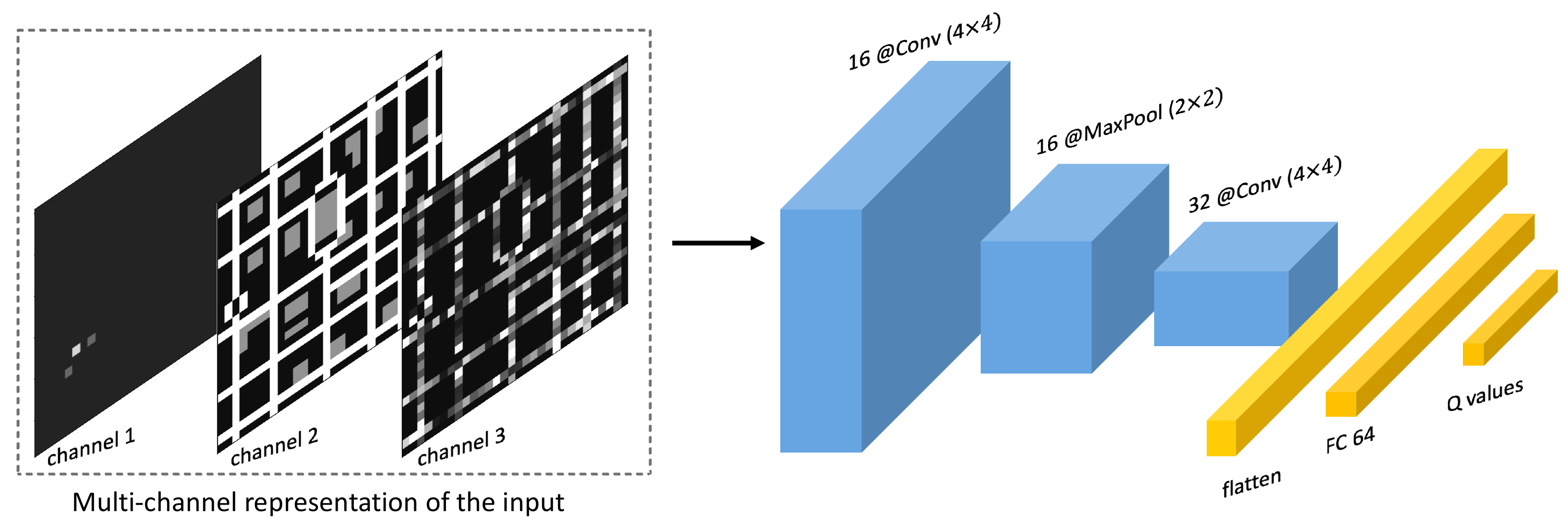}
         \vspace{4.5mm}
         \caption{}
         \label{fig:cnn}
     \end{subfigure}
        \caption{(a) Considered area in down-town Toronto for the multi-robot traffic monitoring scenario. The task of the agents is to monitor the traffic condition on the main roads which are highlighted by the orange color. (b) Multi-channel input and architecture of the considered neural network. The first channel of the input is the position of the ego vehicle and the neighboring vehicles. The second channel is the map of the environment, and the third channel is the local uncertainty map.
 }
        \label{fig:config}
\end{figure*}

\noindent \textbf{Action.} We denote action of the $i$-th agent by $a_i(t)$. At each time $t$, the agent can change its current cell and go to one of its neighboring cells such as north, south, west, east, northwest, northeast, southwest, and southeast. The agent can also remain in its current cell. Hence, the action space of each agent has size $9$.

\noindent \textbf{Reward.}
We consider a reward function that has the following components:
\begin{itemize}
    \item $r_t^{c}$: A negative reward given to each agent if it collides the obstacles (both static and dynamic ones) or goes to a no-fly area. Using this reward, the agents learn to satisfy constraint \eqref{eq:collision_free}.
    \item $r_t^{n}$: A positive reward given to an agent if visits a cell that has not been visited so far.
    \item $r_t^{u}$: A positive reward given to an agent to motivate it to visit locations with higher uncertainties. Since the agents' goal is to minimize the average uncertainty of the road network, a good strategy for the agents is to visit the locations with higher uncertainties, as the contribution of such locations in \eqref{eq:avg_uncertainty} is more than locations with small uncertainties. To achieve this goal, we use the uncertainty function $u_k (t)$ to define $r_t^{u}$ as $r_t^{u} = u_k (t)$.
\end{itemize}
Given these sub-rewards, the reward function is defined as 
\begin{equation}
    \label{eq:reward}
    r_t = r_t^{c} + r_t^{n} + \lambda r_t^{u}, 
\end{equation}
where $\lambda$ is the parameter of the reward function.

\subsection{Algorithm Description}
To solve the given POMDP, we use a distributed algorithm based on DQN. The description of the algorithm is given in Algorithm \ref{alg:distributedDQN}. At each episode, we randomly initialize positions of the agents in environment. At each time, the agents adopt $\epsilon$-greedy policies. Upon taking an action, each agent stores its experience tuple in its local memory. Moreover, the agent updates its local uncertainty map to use in the next round. As described earlier, the information exchange between the agents and the TMC takes place every $T_u$ time units. In this stage, the agents send their visited locations and the corresponding experience tuples (stored in the local memories) to the TMC. The TMC adds these experiences to the global memory $\mathcal{D}$. The TMC uses the received data to update the global uncertainty map and sends this map to all agents. To train and update the parameter of the Q-network at each step of the episodes, a mini-batch of size $b$ is sampled from $\mathcal{D}$ and the corresponding loss in \eqref{eq:loss} is minimized. This can be carried out in either centralized or decentralized fashion. The procedure continues until the Q-network is trained.

\begin{figure*}
     \centering
     \begin{subfigure}{.23\textwidth}
         \centering
         \includegraphics[width=1\textwidth]{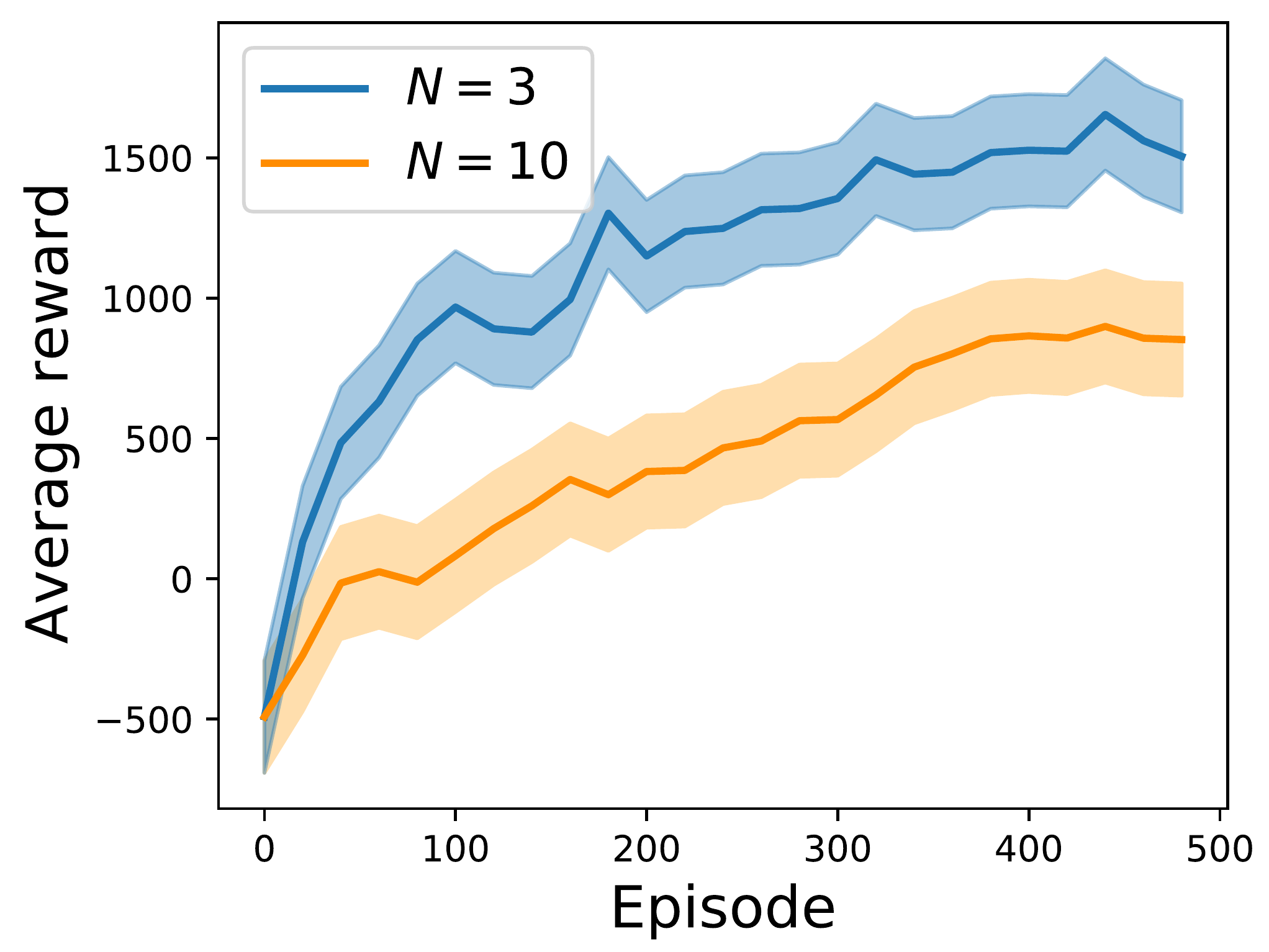}
         \caption{}
         \label{fig:Reward_s1}
     \end{subfigure}
     \begin{subfigure}{.23\textwidth}
         \centering
         \includegraphics[width=1\textwidth]{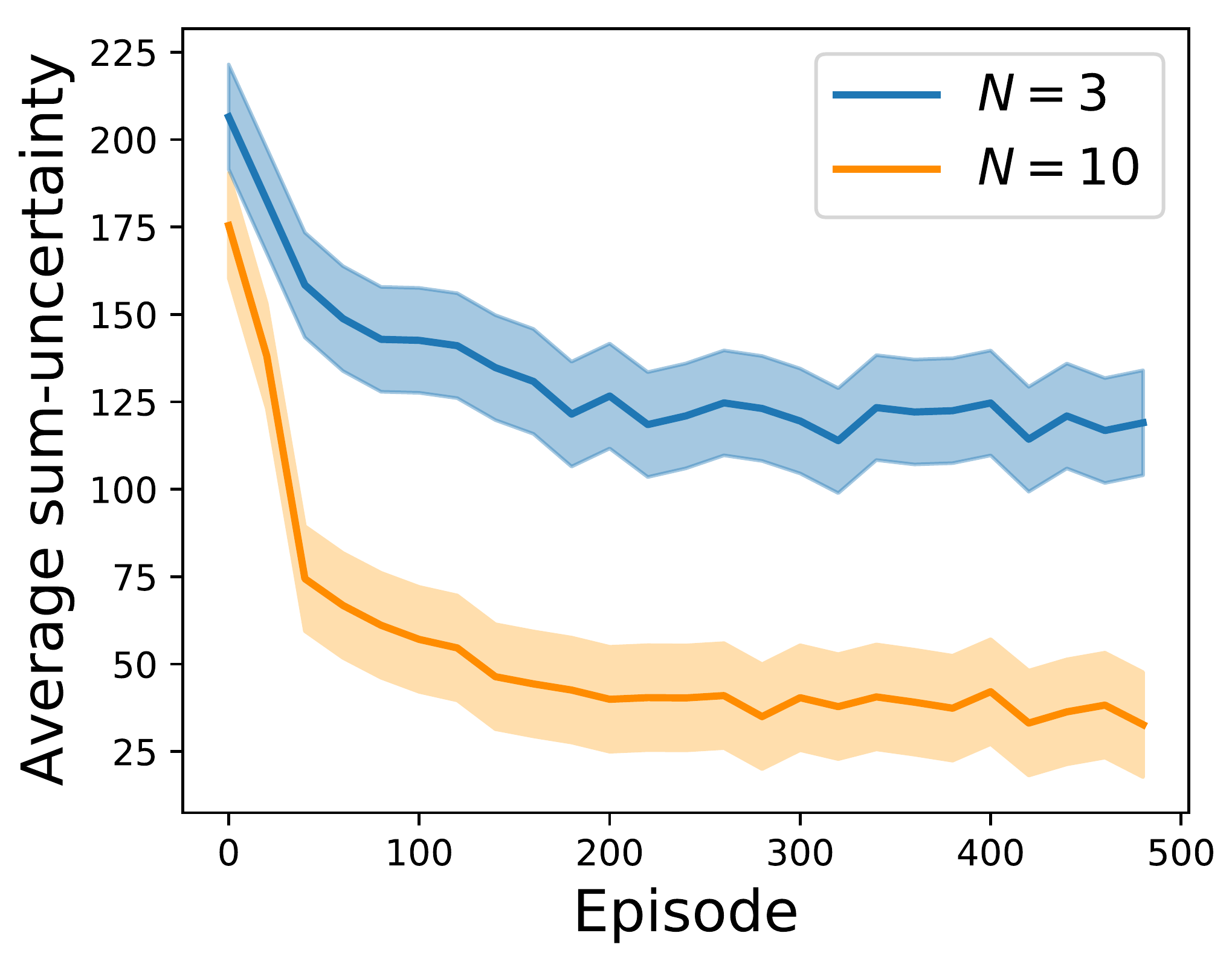}
         \caption{}
         \label{fig:Uncertainty_s1}
     \end{subfigure}
     \begin{subfigure}{.23\textwidth}
         \centering
         \includegraphics[width=1\textwidth]{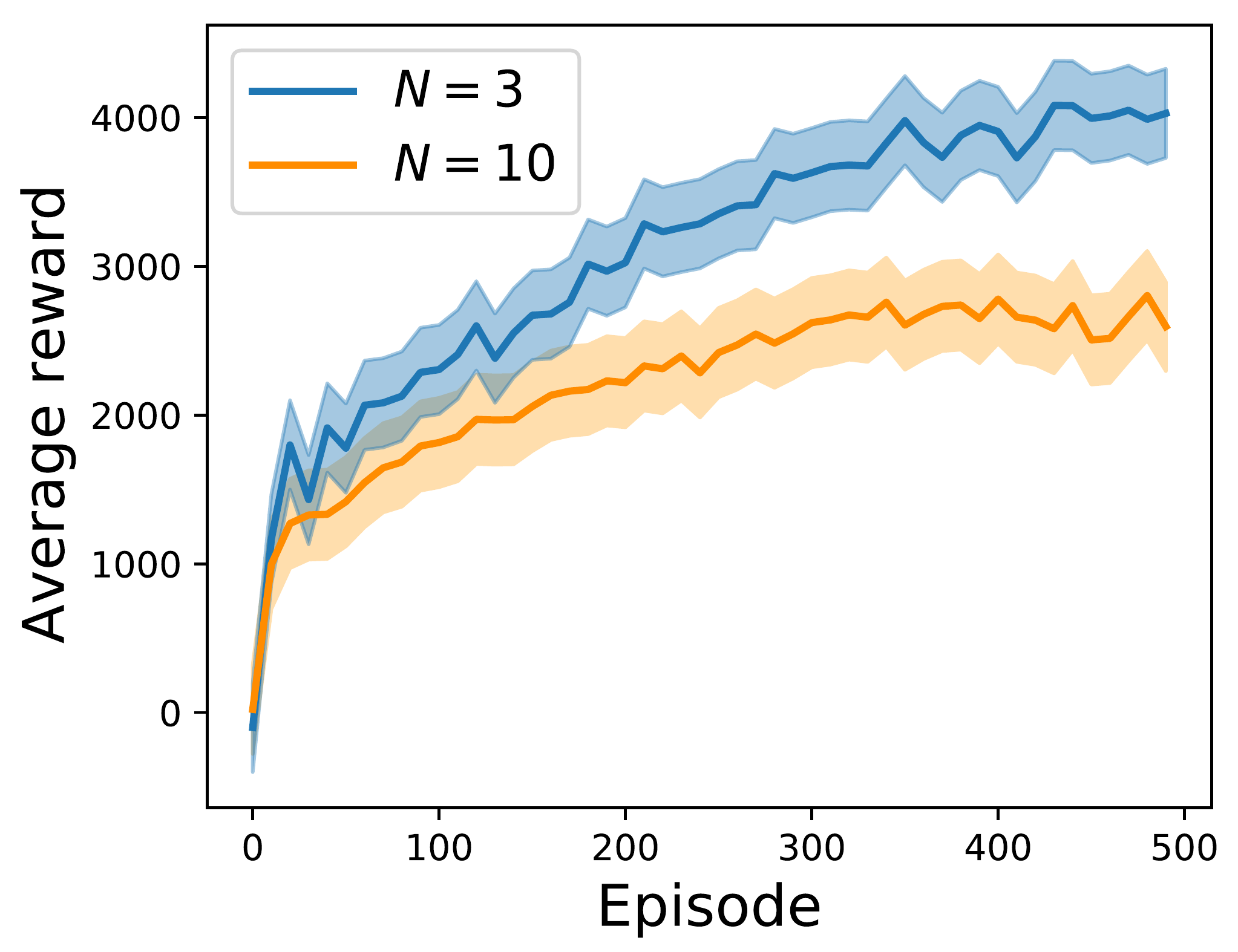}
         \caption{}
         \label{fig:Reward_s2}
     \end{subfigure}
     \begin{subfigure}{.23\textwidth}
         \centering
         \includegraphics[width=1\textwidth]{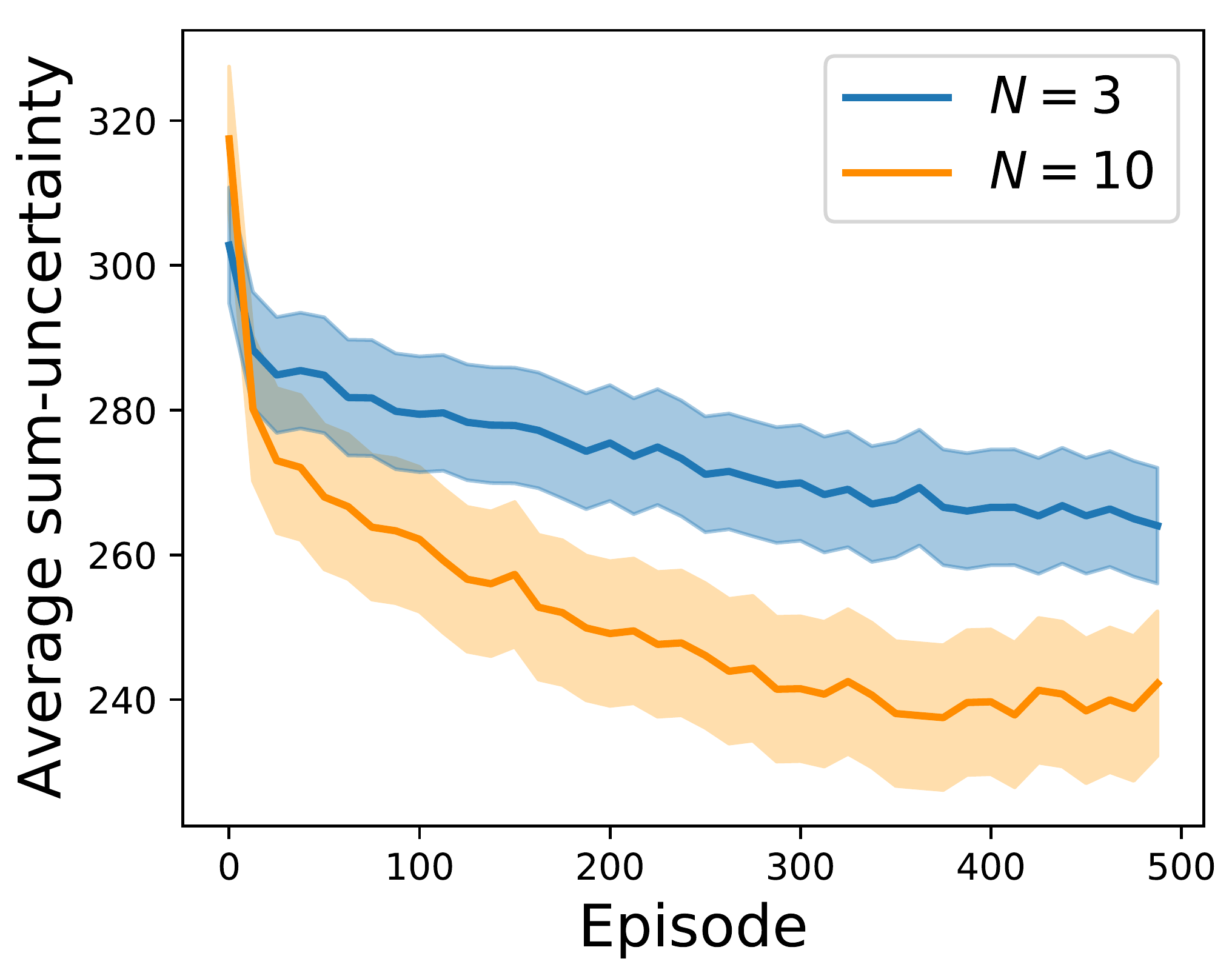}
         \caption{}
         \label{fig:Uncertainty_s2}
     \end{subfigure}
        \caption{The training curves of our algorithm. (a) and (b) correspond to \textit{scenario I} and (c) and (d) are for \textit{scenario II}.}
        \label{fig:training}
\end{figure*}

\section{Evaluation and Results}

In this section, we evaluate the performance of our proposed algorithm on a real road network topology and present the results.

\subsection{Experimental Setup and Environment}
We implement the multi-robot traffic monitoring environment in Python. For our environment, we consider an area in downtown Toronto, as shown in Fig. \ref{fig:Toronto}. The task of the aerial vehicles is to monitor the traffic condition on the main roads. We represent the given area with a grid of size $30 \times 30$. We limit the maximum speed of the aerial vehicles to $2\frac{m}{s}$ to allow them to capture precise images of the traffic condition. The UAV's actions are made once per minute. Hence, each time slot in our evaluation is $1$ min. We assume that $\alpha_k = \alpha, \forall k$. Unless otherwise stated, the value of $\alpha$ is set to $0.01$.

\begin{table}[t]
    \centering
        \caption{Hyper-parameters used for the training.}
        \begin{tabular}{ l  l }
        \noalign{\hrule height 0.05cm}
            Parameter & Value \\
            \hline 
            Adam optimizer learning rate &  $0.001$\\
            replay memory size &  $100000$\\
            mini-batch size ($b$)&  $128$\\
            target network update frequency ($f$) & every $5$ episodes\\
            discount factor ($\gamma$) & $0.99$ \\
            filter size of the convolutional layers & $(4 \times 4)$\\
            size of fully-connected layer & $64$\\
            number of training episodes ($E$) &  $500$\\
            maximum number of steps per episode ($T_{ep}$) & $1000$\\
            decaying for the $\epsilon$-greedy algorithm & $0.5$ to $0.05$\\
            \noalign{\hrule height 0.05cm}
        \end{tabular}
    \label{table:hyper_parameters}
\end{table}

\subsection{Neural Network and Implementation Parameters}
The structure of the neural network is given in Fig. \ref{fig:cnn}. The network has both convolutional (Conv) and fully connected (FC) layers. The size of the input image is $30 \times 30 \time 3$. We use rectified linear unit (ReLU) function as the activation function for both convolutional and fully connected layers. The last layer of the network has no activation as it estimates the Q-function values. For training, each agent implements an $\epsilon$-greedy policy to select its action. The value of $\epsilon$ is gradually annealed from $0.5$ to $0.05$. The parameters used for the training are given in Table \ref{table:hyper_parameters}. The sensing range of the agents ($r_s$) is set to $1.5d$, where $d$ the width of each grid cell. In our experiment, $d=60$m. For the reward function in \eqref{eq:reward}, we consider the following components: $r_t^c=-20$, $r_t^n = +1$, and $\lambda = 5$.

\subsection{Results and Discussion}

Fig. \ref{fig:training} shows the training curves of our algorithm for both scenarios. The number of agents are considered as $N=3$ and $N=10$. At the beginning of the training, the agents do not know the optimal policy. Hence, they take inefficient actions. As training continues, the agents learn to adopt their paths such that the average uncertainty in the environment decreases. In \textit{scenario I}, the agents learn to maximize their visits to locations with active events. However, in \textit{scenario II}, they learn to visit locations with high uncertainty values more frequently. We also observe that the average reward decreases with the number of agents. In fact, according to \eqref{eq:reward}, the received reward of each agent depends on the uncertainty of the visited location. As the number of agents increases, the uncertainty of the network decreases since we have more monitoring resources. Accordingly, the uncertainty term in \eqref{eq:reward} will have a smaller value which in turn reduces the received reward. 

\begin{figure}
     \centering
     \begin{subfigure}{.24\textwidth}
         \centering
         \includegraphics[width=1\textwidth]{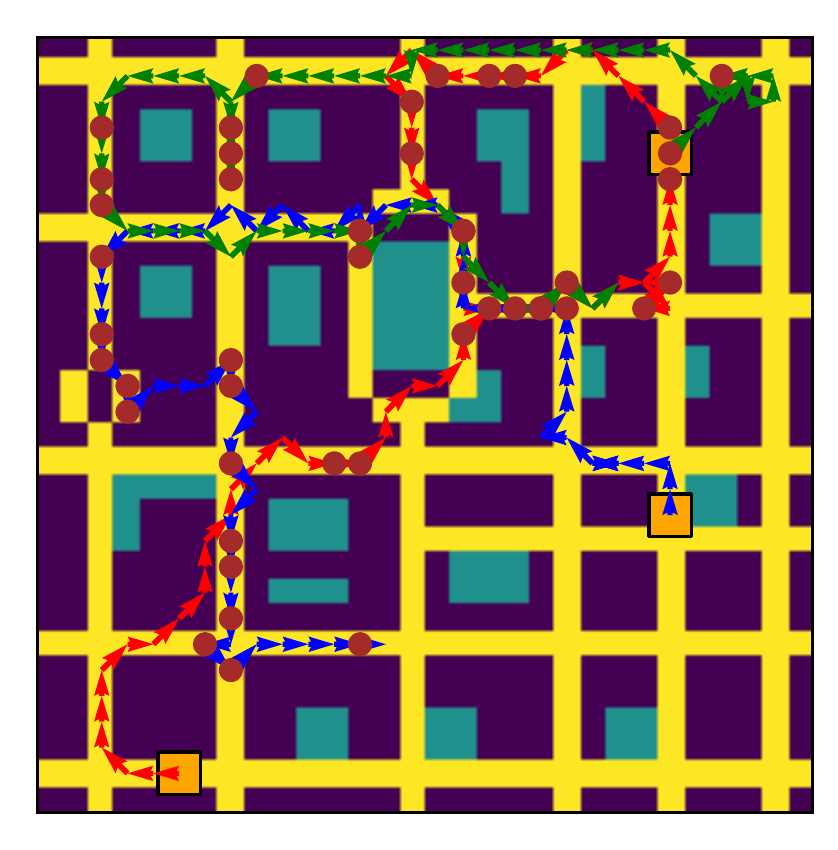}
         \caption{}
         \label{fig:path_s1}
     \end{subfigure}
     \begin{subfigure}{.24\textwidth}
         \centering
         \includegraphics[width=1\textwidth]{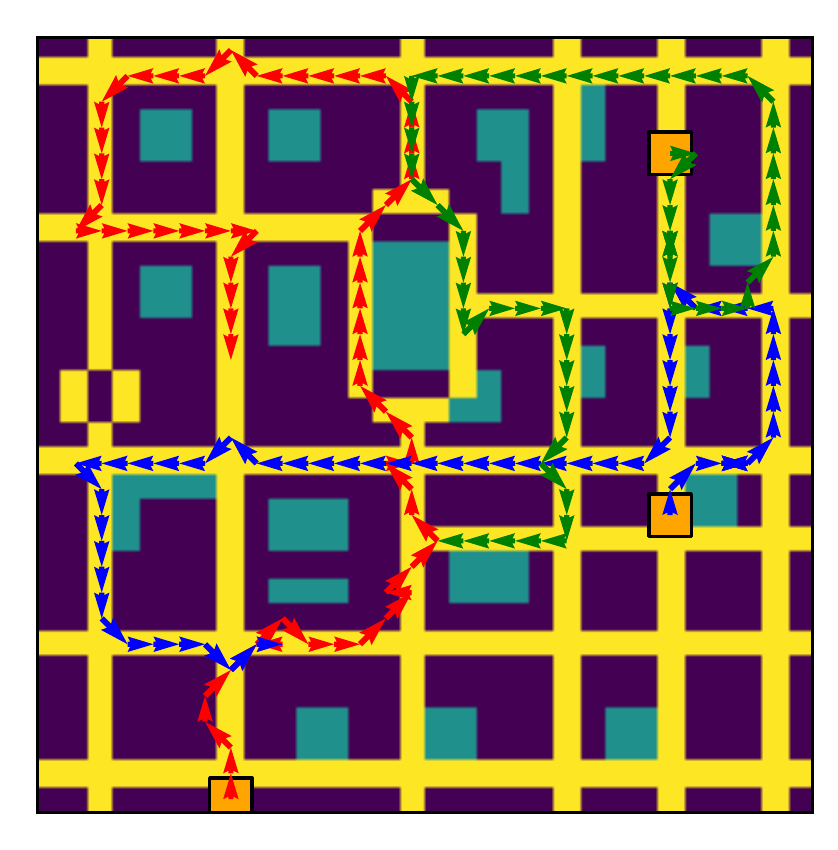}
         \caption{}
         \label{fig:path_s2}
     \end{subfigure}
     \begin{subfigure}{0.5\textwidth}
         \centering
         \includegraphics[width=0.95\textwidth]{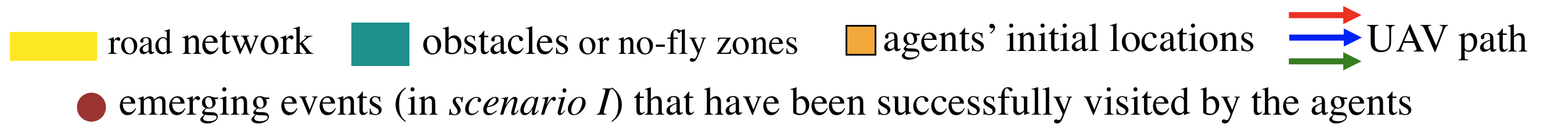}
     \end{subfigure}
        \caption{Sample paths of the aerial vehicles for (a) \textit{scenario I} and (b) \textit{scenario II}.}
        \label{fig:path}
        \vspace{-0.1cm}
\end{figure}

Fig. \ref{fig:path} shows sample paths for a team of $3$ aerial vehicles in both scenarios. In Fig. \ref{fig:path_s1}, we assume that there is no active event at the beginning, and all events emerge during the monitoring period. In contrast, in Fig. \ref{fig:path_s2}, we consider a random uncertainty map at the beginning of the monitoring cycle. The task of the aerial vehicles in both scenarios is to dynamically adjust their paths to minimize the uncertainty in the network. In Fig. \ref{fig:path_s1}, the agents learn to successfully visit the traffic events that emerged during the mission period to capture real-time images. However, in Fig.  \ref{fig:path_s2}, the agents choose their paths to visit locations with a high uncertainty value. Moreover, we observe that aerial vehicles learn to fly over the roads almost all time instances. Even for changing the roads, instead of choosing the shortest paths, they pick longer paths that cover the roads. 

\begin{figure}
     \centering
     \begin{subfigure}{.22\textwidth}
         \centering
         \includegraphics[width=1\textwidth]{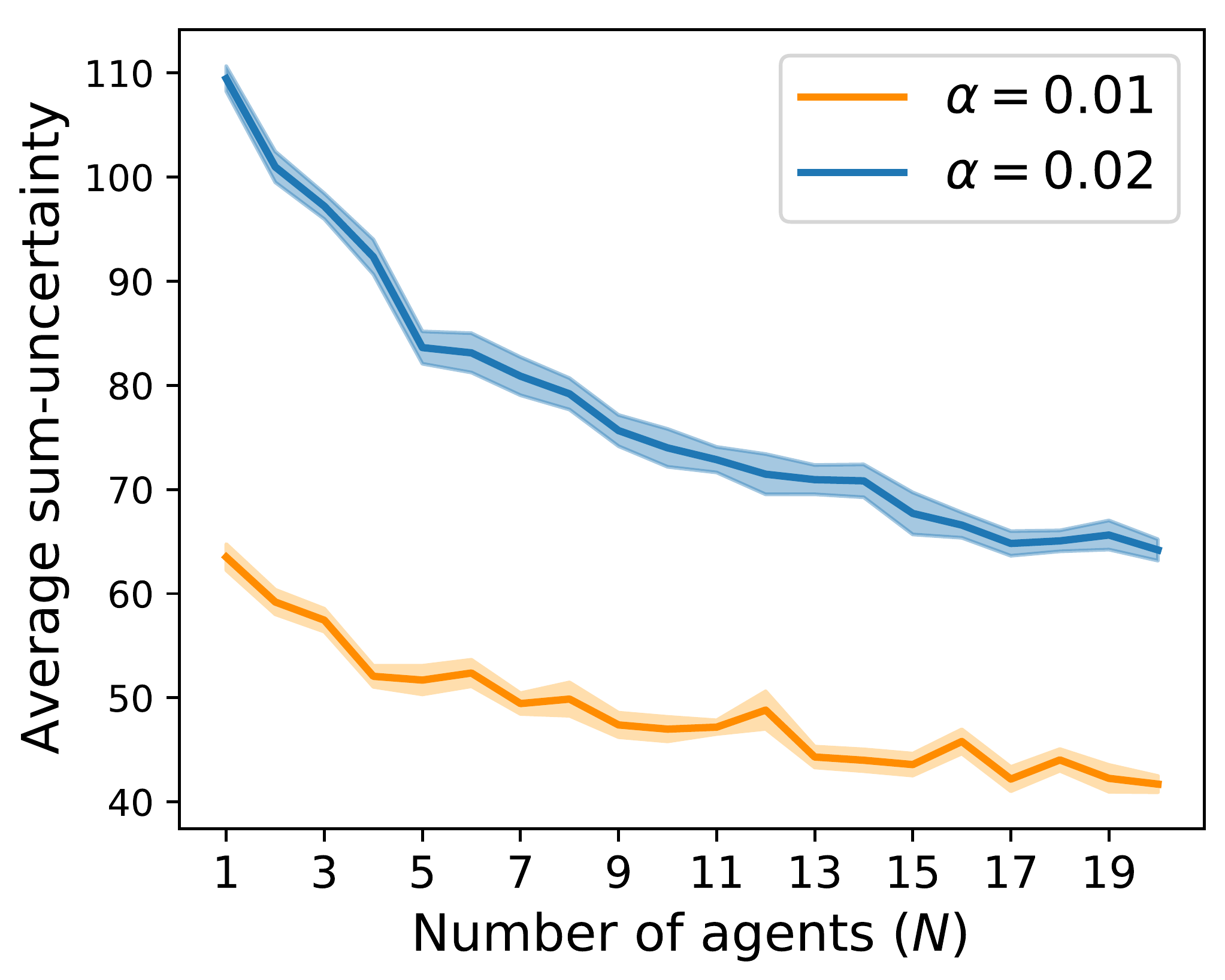}
         \caption{}
         \label{fig:uncertainty:uncertainty1}
     \end{subfigure}
     \begin{subfigure}{.22\textwidth}
         \centering
         \includegraphics[width=1\textwidth]{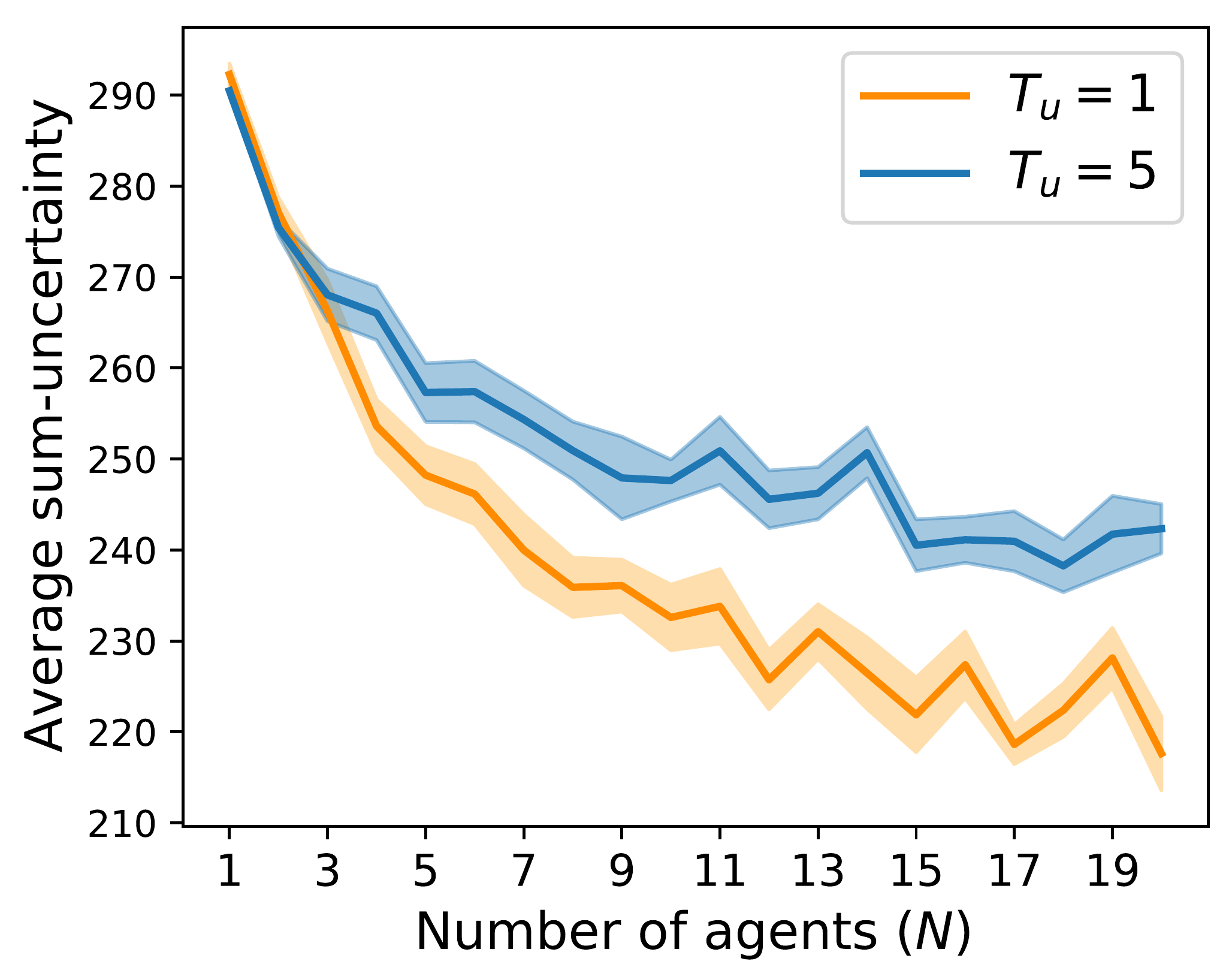}
         \caption{}
         \label{fig:uncertainty:uncertainty2}
     \end{subfigure}
        \caption{Average uncertainty of the road network in (a) \textit{scenario I} and (b) \textit{scenario II}.}
        \label{fig:uncertainty}
\end{figure}

The number of agents ($N$) is another important factor that affects performance of the traffic monitoring system. In Fig. \ref{fig:uncertainty}, we present the uncertainty of the road network for both scenarios as a function of $N$. As we expect, the average uncertainty decreases with the number of agents. This reduction is more significant in \textit{scenario I} compared to \textit{scenario II}. The reason for this comes from the difference between the uncertainty models in \eqref{eq:uncertainty_1} and \eqref{eq:uncertainty_2}. In \textit{scenario I}, after visiting a location with an active event, the uncertainty of that location will be $0$. This value remains unchanged until another event emerges in the mentioned location. In contrast, in \textit{scenario II}, after visiting a location, its uncertainty does not remain constant. In other words, the uncertainty value is set to $0$ upon the visit. However, the uncertainty value increases as time passes (see equation \eqref{eq:uncertainty_2}). Hence, the value of uncertainty will be higher in this scenario, and accordingly, the percentage of the uncertainty reduction will be smaller.

Fig. \ref{fig:uncertainty:uncertainty2} shows the effect of $T_u$ on the performance of the proposed algorithm. When $T_u$ is small, the agents will communicate with the TMC more frequently ($T_u=1$ corresponds to the continuous communication between the agents and the TMC). As a result, their uncertainty models will be more accurate than when the agents communicate less often. This accuracy improves the probability that the agents visit locations with higher uncertainty values. We explain this issue with one example. Consider cell $k$ with a high uncertainty value and assume that this cell has been visited by agent $i_1$ at time $t$. Moreover, assume that agent $i_1$ is not in $\mathcal{N}_{i_2} (t)$. The true uncertainty of this location will be $0$ after the visit. However, since this cell is not in the sensing range of agent $i_2$, agent $i_2$ does not become aware of the visit. Accordingly, agent $i_2$ does not update the uncertainty of cell $k$ in its local uncertainty map. In contrast, it considers cell $k$ as a location with a high uncertainty value. Accordingly, the agent considers cell $k$ as a candidate for its next visit. The number of these inefficient visits increases by the value of $T_u$. It is worth mentioning that the resulting gap in uncertainty values will be negligible for a small number of agents. However, as the number of agents increases, it becomes more important to have precise knowledge of the environment to make efficient decisions.

\section{Conclusions}

We studied the traffic monitoring problem in a road network using a fleet of UAVs. To address the stochastic nature of the traffic events, we used an uncertainty metric to model the traffic monitoring problem. We considered two different scenarios, depending on the communication mode between the agents and the TMC. In the first scenario, we assumed that the agents continuously exchange information with the TMC, and hence, they have complete and real-time knowledge of the environment. However, in the second scenario, we assumed that the communication between the agents and the TMC is limited to specific time instances. Moreover, the observation of each agent is restricted to its sensing range. Therefore, the agents have partial observation of the environment. To develop a framework that works in both cases, we expressed the traffic monitoring problem as a POMDP and proposed a distributed algorithm based on deep Q-learning to control the agents' movements. Experimental results showed the effectiveness of our proposed algorithm in reducing uncertainty of the environment.

\bibliographystyle{IEEEtran}
\bibliography{bibfile}

\end{document}